\documentclass{article}
\usepackage{ijcai20}

\usepackage{times}
\usepackage{soul}
\usepackage{url}
\usepackage[hidelinks]{hyperref}
\usepackage[utf8]{inputenc}
\usepackage[small]{caption}
\usepackage{graphicx}
\usepackage{amsmath}
\usepackage{amsthm}
\usepackage{booktabs}
\usepackage{algorithm}
\usepackage{breqn}
\usepackage{latexsym}
\usepackage{amsfonts}
\usepackage{setspace}
\usepackage{multirow}
\usepackage{algpseudocode} 
\usepackage{amssymb}
\title{TVDIM: Enhancing Image Self-Supervised Pretraining \\
via Noisy Text Data}

\author{
Pengda Qin$^1$
\and
Yuhong Li$^1$
\and
Kefeng Deng$^1$
\and
Qiang Wu$^1$
\affiliations
$^1$Alibaba Group
\emails
\{pengda.qpd, daniel.lyh, kefeng.deng, qiangwu.wq\}@alibaba-inc.com
}
\begin{document}

\maketitle

\begin{abstract}
Among the multi-modal signals of the real world, language is the modality nearest to human understanding level; in contrast, vision reflects the real world honestly and objectively. When in a visual scene, machines are expected to understand visual information at the human level.
Inspired by this, we propose a novel self-supervised visual learning method, named \textbf{T}ext-enhanced \textbf{V}isual \textbf{D}eep \textbf{I}nfo\textbf{M}ax (\textbf{TVDIM}) strategy, to learn visual representations enhanced by language-modality information. The co-occurrence of textual and visual information commonly exists around us, such as books and internet; More importantly, this text provides the annotation-free information for better understanding its corresponding images. Based on that, we make full use of this information overlap, and utilize the concept of mutual information maximization to integrate textual information into visual representations in a self-supervised way. Considering the information gap between different modalities, we adopt contrastive learning to measure mutual information by \emph{Ranking}. During evaluation, we directly use the pretrained visual representations to complete various image classification tasks. Experimental results show that, with the assistance of annotation-free text information, TVDIM significantly outperforms previous visual self-supervised methods by processing the same set of images.
\end{abstract}

\section{Introduction}
\label{introduction}

\begin{figure}[t]
	\vspace{-2mm}
	\center
	\includegraphics[width=7.2cm]{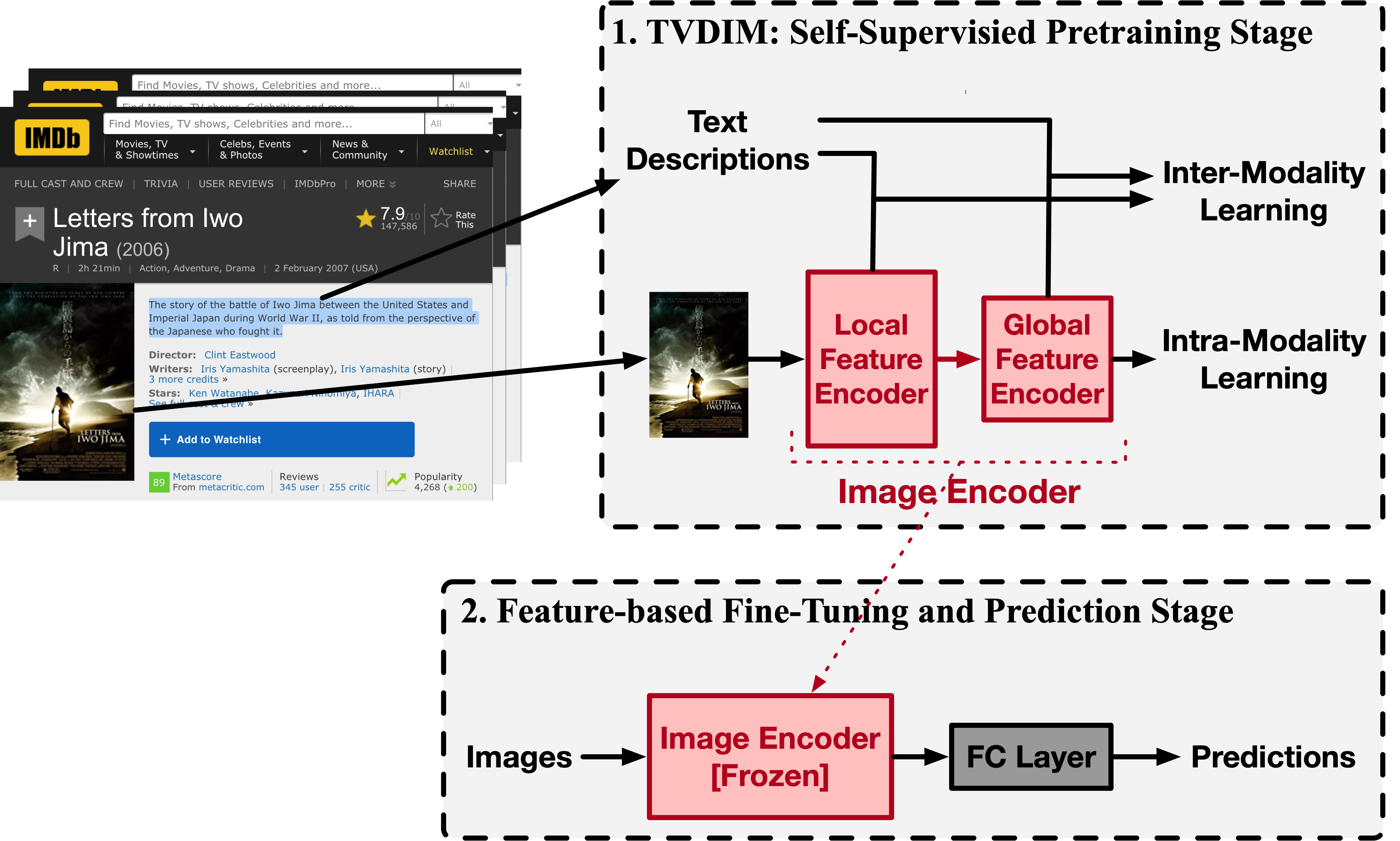}
	\vspace{-2mm}
	\caption{Our overall pipeline consists of two stages. The first stage is to do self-supervised training for image encoder via the proposed TVDIM. Noted that the training image-text pairs naturally exist in various medias and can be obtained without human annotation. The second stage is to freeze image encoder and directly do feature-based fine-tuning and prediction for downstream tasks.}
	\label{fig:ex-intro}
	\vspace{-2mm}
\end{figure}

Human access the real world via multi-modal signals everytime and everywhere. Most part is the original signal that is equivalent to all living things in this real world, like vision and sound, while a small part is unique to humans. Language is obviously in the latter camp, which is the product of human evolution, and it is ever-evolving. 
Therefore, compared with vision, the information expressed by language is quite closer to be the human understanding level. Not only that, theoretically a visual scene can provide infinite information along with the unlimited increasing of resolution, while human just selectively receive information to make reasonable reactions. 
%Human's selection ability is inherent, which is derived from human cognitive ability.
To promote machine intelligence to human level, many advanced architectures have been proposed and their excellent learning ability has been fully proved. Deep neural networks, especially convolutional neural networks (CNNs), have drastically advanced the development of vision-related tasks. The conventional way is to train deep neural networks in a fully supervised manner; However, not only it is prohibitively expensive due to a large amount of efforts on manual labeling, but also the randomly-initialized parameters violate human learning process and can not provide any prior valuable knowledge for the following tasks. Under this background, self-supervised learning (SSL) is a rational solution to overcome these issues. As for visual representation learning, the typical paradigm is to pretrain a large model via well-designed pretext tasks from a large annotation-free dataset. The definitions of pretext tasks in previous works are devoted to exploring rational supervised signals from the image spatial structure \cite{doersch2015unsupervised,noroozi2016unsupervised}, multi-view of image \cite{tian2019contrastive} or relationships between local and global representations \cite{bachman2019learning}. We believe that these vision-level pretext tasks can encode useful prior knowledge, such as color, position and spatial structure; However, we argue that it merely stays at vision-level (\emph{intra-modality learning}), and \emph{inter-modality} prior knowledge is also conducive to the performance of various visual tasks.

In this paper, we present a novel image self-supervised pretraining strategy TMDIM that considers both the \emph{intra-modality} representation learning and \emph{inter-modality} representation learning. The intra-modality component focuses on modeling the vision-level knowledge, where we adopt the recently proposed method of Augmented Multiscale Deep InfoMax (AMDIM) \cite{bachman2019learning}. The pretext task of AMDIM is to maximize mutual information between a global summary feature vector and a collection of local feature vectors pulled from intermediate layers. Noise-contrastive estimation (NCE) \cite{gutmann2010noise,mcallester2018formal,poole2019variational} is used to model mutual information maximization. The inter-modality component integrates the image-related text to enhance image encoder at the human understanding level. Instead of predictive learning \cite{joulin2016learning,li2017learning}, we adopt contrastive learning \cite{tian2019contrastive} to realize mutual information maximization between image-text pairs from both global and local feature level. It is worthy of note that, in this inter-modality case, NCE just yields the suboptimal performance, because the image-text pairs exist \textbf{information gap} even though their information is highly matched. To solve this problem, we decide to model the inter-modality mutual information maximization in a pairwise ranking way \cite{kiros2014unifying}. Unlike the infinite penalty of NCE loss function, pairwise ranking loss function defines a margin value to limit the boundary of penalty level, and our experimental results validate that pairwise ranking loss function achieves better performance and makes the training process more stable. Besides the superiorities mentioned above, the integration of inter-modality learning increases the difficulty of self-supervised learning, which is also beneficial to more robust and generic visual representations. 
Our main contributions are three-fold:
\begin{itemize}
\item TVDIM is the first to consider both intra-modality and inter-modality self-supervised learning for visual representation learning;
\item We prove that pairwise ranking loss is more suitable to model inter-modality mutual information maximization;
\item TVDIM is able to enhance visual representations at the human understanding level, and it significantly improves the performance of various downstream tasks.
\end{itemize}

\section{Related Work}
\label{relatedwork}
\begin{figure}[t]
	\vspace{-2mm}
	\center
	\includegraphics[width=7.5cm]{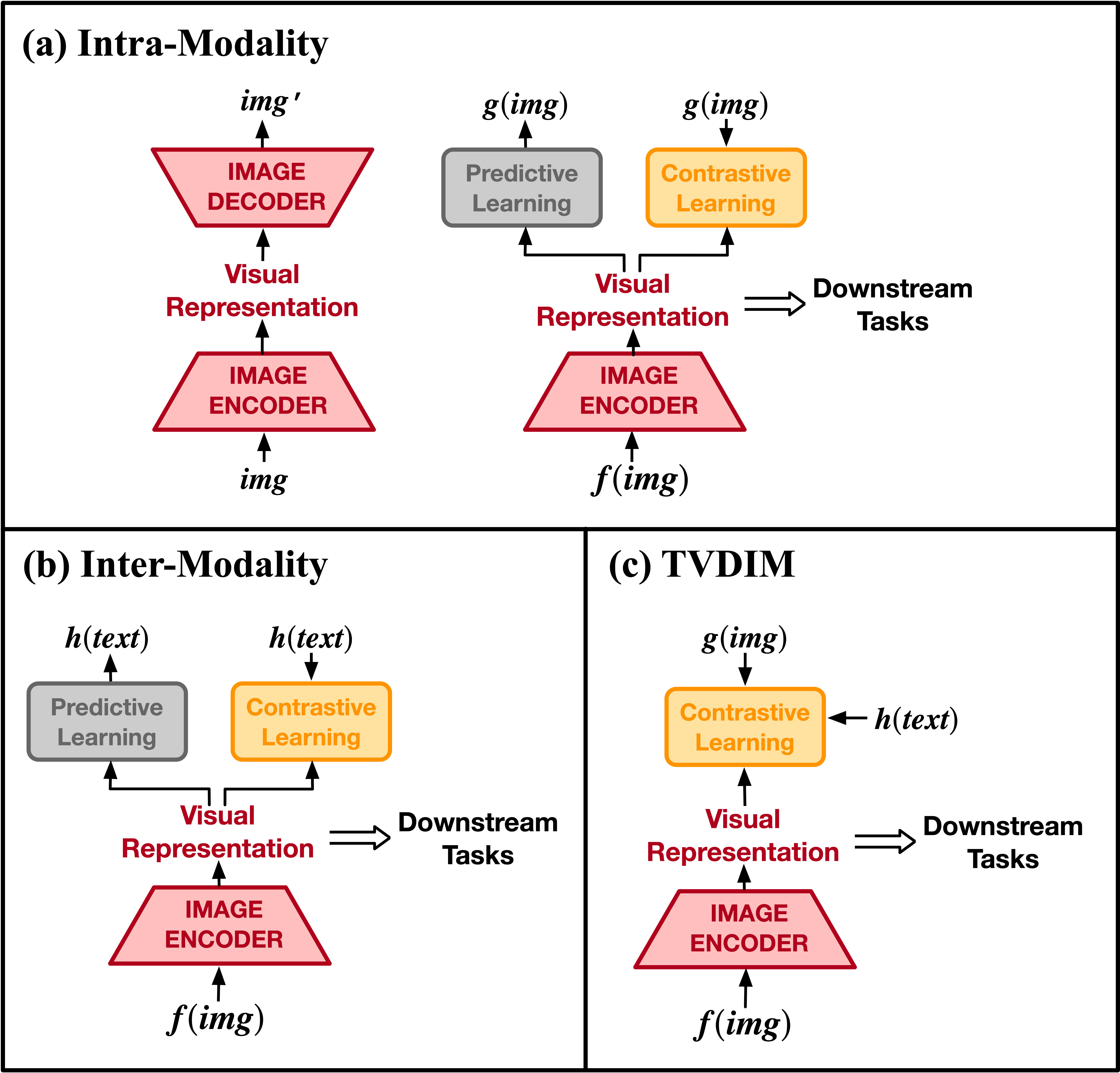}
	\vspace{-2mm}
	\caption{Previous related works can be divided into two camps, intra-modality learning (a) and inter-modality learning (b). Based on them, TVDIM (c) combines their merits to further robust self-supervised visual representation learning.}
	\label{fig:related-works}
	\vspace{-2mm}
\end{figure}

%Unsupervised learning seeks to explore the implicit value of data without human intervention. 
As an important member of unsupervised-learning family, the objective of self-supervised learning is more clear and definite. Self-supervised learning is to make subsequent problem solving easier via learning transformations of the data. In the field of visual representation learning, previous research works are designed either in an intra-modality way or in an inter-modality way.
Figure \ref{fig:related-works}(a) concludes two different kinds of intra-modality learning. The left one introduces a traditional unsupervised learning technique that aims to rebuild inputs from visual representations \cite{rumelhart1985learning,hinton2006reducing,erhan2010does}. However, it is proved that these works just achieve the goal of data compression, and merely a small amount of valuable knowledge for downstream task is learned. To overcome this problem, current works (right part of Figure \ref{fig:related-works}(a)) resort to the reasonable transformations ($f(*)$, $g(*)$, $h(*)$) of data to encode more valuable information without human intervention, including counting the objects in a image \cite{noroozi2017representation}, recovering colors from grayscale images \cite{zhang2016colorful}, or predicting image spatial structure \cite{noroozi2016unsupervised,kim2018learning,doersch2015unsupervised,oord2018representation,gidaris2018unsupervised}. These pretext tasks belong to predictive learning. Recently, contrastive learning is also considered to increase the diversity and difficulty of pretext tasks. 
Contrastive Multiview Coding (CMC) \cite{tian2019contrastive} learns representations that capture information shared between multiple sensory views of an image. Deep InfoMax (DIM) \cite{hjelm2018learning} and AMDIM \cite{bachman2019learning} perform representation learning by maximizing mutual information between feature representations from different encoder layers. Besides that, the concept of inter-modality learning has already been proposed previously (Figure \ref{fig:related-works}(b)). Kiros et al. \shortcite{kiros2014unifying} present a contrastive learning approach for learning visual-semantic embeddings for cross-modality retrieval, and their text part merely consists of words or phrases. Mahajan et al. \shortcite{mahajan2018exploring} utilize the hashtags as the supervised signals and perform the predictive learning on billions of social media images. Different from these works, we are the first work to consider both intra-modality and inter-modality learning for robust visual representations. More than that, we are in a more complicated setting: our image-related text could be a sentence or a paragraph. We believe that, visual intra-modality learning is beneficial to explore more visual information, and vision-text inter-modality learning enables image encoder to better understand visual information at the human understanding level.

\begin{figure*}[t]
\begin{center}
\includegraphics[width=16cm]{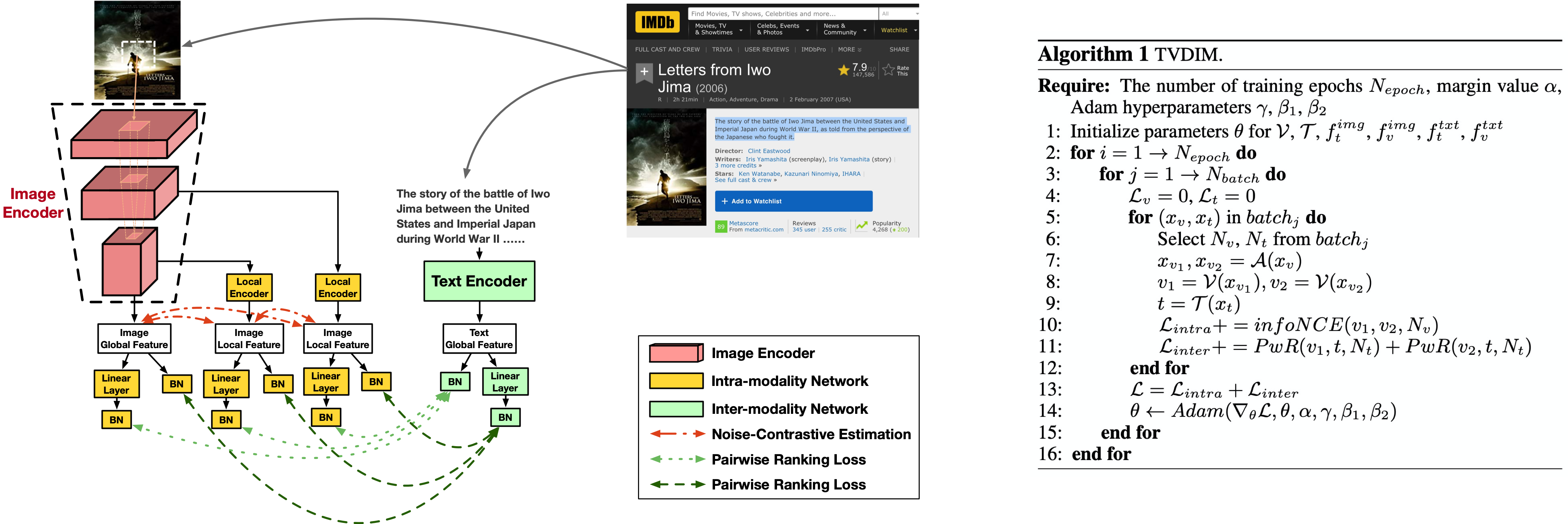}
\caption{Overview of TVDIM. Our self-supervised pretraining consists of intra-modality learning (red arrows) and inter-modality learning (green arrows). Our ultimate goal is to learn a sophisticated image encoder (red part) for better visual representations. In terms of intra-modality, we integrate visual-related prior knowledge into image encoder by maximizing NCE-based infomax from multiple scales. Data augmentation part is not presented here for clarity. In terms of inter-modality, we first encode text into text representations, and then maximize the mutual information between text and image via pairwise ranking. This inter-modality mutual information maximization is calculated in both textual feature space (light green arrows) and visual feature space (dark green arrows).}
\label{fig:framework}
\end{center}
\end{figure*}

Currently, with the significant advance of transformer-based language pretraining \cite{devlin2018bert,yang2019xlnet}, a series of works \cite{tsai2019multimodal,lu2019vilbert,tan2019lxmert,tan2019lxmert,sun2019videobert} aim to do multi-modality self-supervised learning based on transformer network. They all utilize self-attention mechanism \cite{vaswani2017attention} to model intra-modality learning. In the meantime, Tan and Bansal \shortcite{tan2019lxmert} propose a cross-attention mechanism to model inter-modality learning; Lu et al. \shortcite{lu2019vilbert} propose a co-attention mechanism to enable the fusion of cross-modality information. In terms of motivation, these works focus on better multi-modality representation, while TVDIM aims to obtain better visual representations. Besides that, the image input formats are different. TVDIM is based on the original image signals. To satisfy the transformer structure, these works need to preprocess images into a series of ROI features via the pretrained R-CNN \cite{ren2015faster} network; Thus, error propagation is inevitable.

\section{Methodology}
\label{sec:methodology}
The self-supervised training of TVDIM consists of two parallel processes (Figure \ref{fig:framework}), intra-modality learning and inter-modality learning. Accordingly, the following descriptions are divided into two parts. The first part presents a brief introduction of visual intra-modality learning. Given a positive image pair and a set of negative pairs, we aim to do multi-scale infomax across multiple views via NCE loss function. The second part detailedly introduces vision-text inter-modality learning. Given a positive image-text pair and a set of negative image-text pairs, we use pairwise ranking loss to maximize the multi-scale mutual information between the positive pair.

\subsection{Visual Intra-modality Learning}
%We select Augmented Multiscale Deep InfoMax (AMDIM) as the targeted self-supervised visual representation learning model. 
Our intra-modality learning follows the \emph{infomax} idea of DIM and AMDIM. AMDIM is the extended version of DIM. Given an image input $x_v$, the infomax process is implemented within the interior structure of image encoder ${\mathcal{V}}$. 
The core concept is to maximize mutual information between antecedent features $v^a = {\mathcal{V}}_{a}(x_v)$ and consequent features $v_{ij}^c \in \{{{\mathcal{V}}_{c}(x_v)}_{ij}: \forall i, j\}$. Antecedent features are the features that encode the image input to condition on, and consequent features are the features to be predicted. The subscripts $i$ and $j$ index the two spatial dimensions of the array of activations in the intermediate layer of image encoder ${\mathcal{V}}$. AMDIM constructs a distribution $p(v^a, v_{ij}^c)$ over (antecedent, consequent) feature pairs via ancestral sampling: First step is to use a data distribution $\mathcal{D}$ to sample input $x_v \sim \mathcal{D}$, and second is based on two uniform distributions $u(i)$, $u(j)$ to select the valid spatial indices $i$, $j$. Given $p(v^a, v_{ij}^c)$ and marginal distributions $p(v^a)$, $p(v_{ij}^c)$, AMDIM seeks an image encoder $\mathcal{V}$ that maximizes the mutual information $I(v^a, v_{ij}^c)$ in $p(v^a, v_{ij}^c)$. AMDIM equally considers three different scales of encoder layers: layer 1, 5 and 7. Respectively, they have the spatial dimension $1 \times 1$, $5 \times 5$ and $7 \times 7$. As for mutual information maximization, they correspond to three different antecedent-consequent pairs, including 1-to-5, 1-to-7 and 5-to-5.

%${\mathcal{V}}_{1} \to {{\mathcal{V}}_{5}(x_v)}$, ${\mathcal{V}}_{1} \to {{\mathcal{V}}_{7}(x_v)}$ and ${\mathcal{V}}_{5} \to {{\mathcal{V}}_{5}(x_v)}$.

Various strategies can be used to realize mutual information maximization. DIM mainly compares two different ways, Jensen-Shannon Divergence (JSD) \cite{lin1991divergence} and Noise-Contrastive Estimation. Their experimental results suggest that infoNCE \cite{oord2018representation} achieves better performance on downstream tasks in most cases, especially when training on less challenging data. 
On balance, both DIM and AMDIM regard infoNCE as the optimal loss function. Noted that AMDIM adopts data augmentation technique \cite{lim2019fast} to further increase the difficulty of of self-supervised learning. During each computation of infoNCE loss, antecedent features $v^a$ and consequent features $v_{ij}^c$ are from augmented views of each input. For a image input $x_v$, AMDIM samples a pair of augmented images $x_v^1 \sim \mathcal{A}(x_v)$ and $x_v^2 \sim \mathcal{A}(x_v)$. Here, $\mathcal{A}(x)$ denotes a distribution of images generated by applying stochastic data augmentation. Therefore, for one kind of antecedent-consequent pairs, the infoNCE-based infomax objective can be formulated as below:
\begin{equation}
\begin{split}
\mathcal{L}_{intra} = \sum_{x_v}\sum_i\sum_j \mathcal{L}_\Phi({\mathcal{V}}_{a}(x_v^1), {{\mathcal{V}}_{c}(x_v^2)}_{ij}, N_{c})
\label{equ:Loss Function}
\end{split}
\end{equation}
where $N_{c}$ denotes the negative samples. $N_{c}$ acts as many distractor consequent features drawn independently from the marginal distribution $p_{\mathcal{A}}({{\mathcal{V}}_{c}(x_v^2)}_{ij})$. Here, $\mathcal{L}_\Phi(*)$ denotes the infoNCE which is a softmax-based version of NCE. 
For brevity, we omit the data augmentation symbols and spatial indices, then the overall infoNCE loss function can be rewrite as follows:
\begin{equation}
\begin{split}
\mathcal{L}_{intra} &= \sum_{v}\sum_i\sum_j \mathcal{L}_\Phi(v^a, v^c, N_{c}) \\
 \mathcal{L}_\Phi(v^a, v^c, N_{c})
 &= -log \frac{exp(\Phi(v^a, v^c))}{\sum_{\tilde{v}^c \in N_{c} \cup  \{v^c\}}exp(\Phi(v^a, \tilde{v}^c ))},
\label{equ:infoNCE}
\end{split}
\end{equation}
%where we omit spatial indices and dependence on $x$ for brevity. 
Empirically, the matching score $\Phi(*)$ is calculated by dot product operation. 

\subsection{Text-Enhanced Visual Inter-Modality Learning}

Simply, self-supervised learning is to find annotation-free supervised signals to do supervised learning. The above subsection merely focuses on the intra-modality signals, while we believe that there exist massive valuable self-supervised signals of inter-modality and this type of information is able to comprehensively enrich visual representations. Compared with vision modality, language modality can express higher-level information and is the nearest modality to human understanding level. Self-supervised learning is to promote downstream tasks, and downstream tasks are defined by human and at the service of human. Therefore, for the purpose of enhancing visual representation learning, we consider language modality as the auxiliary modality. Text is one of the core formats of language modality, and the co-occurrence of text and image is common in various media sources. More importantly, the prerequisite of co-occurrence is the existence of high information overlap between two modalities. This overlapping information provides sufficient conditions to produce rational supervised signals for self-supervised learning. 

Figure \ref{fig:framework} presents the framework we use to enrich visual representations via textual information. We first utilize a text encoder $\mathcal{T}$ to model text information. Due to the role of auxiliary modality, we just consider the text global features $t = \mathcal{T}(x_t)$. On the other side, we still utilize both global visual features and local visual features; For brevity, they are uniformly denoted as $v$. 
%Considering data augmentation, they are uniformly denoted as $v_1$, $v_2$. 
The prerequisite of mutual information calculation is the same feature space. In other words, $t$ and $v$ should be projected into the same unified space. We consider two different spaces: the image feature space and the text feature space. To calculate mutual information in the image space, we adopt a feed-forward layer $f_t^{img}(t)$ which consists of a fully-connected layer and a batch-normalization layer; For $v$, we directly use a batch-normalization layer $f_v^{img}(v)$. As for the text feature space, there are opposite settings: $f_t^{txt}(t)$ is a batch-normalization layer and $f_v^{txt}(v)$ includes a fully-connected layer and a batch-normalization layer. Under the same feature space, we leverage the dot product operation to calculate the matching scores:

%\Phi^{img}(t, v) &\triangleq {f_t^{img}(t)}^{\intercal}f_v^{img}(v)\\
%\Phi^{txt}(t, v) &\triangleq {f_t^{txt}(t)}^{\intercal}f_v^{txt}(v)

\begin{equation}
    \begin{split}
    \Phi^{img}(t, v) &= {f_t^{img}(t)}^{\top}f_v^{img}(v)\\
    \Phi^{txt}(t, v) &= {f_t^{txt}(t)}^{\top}f_v^{txt}(v)
    \label{equ:score}
    \end{split}
\end{equation}
Our experimental results demonstrate that batch normalization plays a crucial role in multimodal mutual information calculation. 

Even though infoNCE is regarded as a good choice for visual intra-modality learning, we argue that there are several drawbacks to apply infoNCE in our inter-modality learning. InfoNCE is formulated in the softmax-based version, which means that the optimal situation is that the probability of positive sample is 1. In other words, infoNCE imposes infinite rewards to positive samples and infinite penalty to negative samples. However, the information overlap between image and text is not equal to perfect alignment. Thus, a certain part of textual information is noise information. Our experimental results also demonstrate that, the performance of infoNCE begins to get the downward trend when the training time exceeds a certain value. To overcome this problem, we decide to substitute it with pairwise ranking ($PwR$) loss. Pairwise ranking loss adopts a margin to define the gap between positive samples and negative samples. When the difference of mutual information is larger than this predefined margin, no penalty will be imposed. Based on that, our pairwise ranking loss is formulated as follows:
\begin{equation}
\begin{split}
\mathcal{L}_{t \to v} = \sum_{v}\sum_k max\{0, \alpha - \Phi(t, v) + \Phi(t, v_k)\}, \\
\mathcal{L}_{v \to t} = \sum_{v}\sum_k max\{0, \alpha - \Phi(v, t) + \Phi(v, t_k)\},
\label{equ:pwr_space}
\end{split}
\end{equation}
where $k$ denotes the index of negative samples, and we omit the calculations in different feature spaces for brevity. To sum up, the overall inter-modality loss function is presented below:
\begin{equation}
\begin{split}
\mathcal{L}_{inter} = \mathcal{L}_{t \to v} + \mathcal{L}_{v \to t} ,
\label{equ:pwr}
\end{split}
\end{equation}

%\begin{algorithm}[t]  
%\small
%\caption{\small{TVDIM.}}  
%\begin{algorithmic}[1]  
%    \Require The number of training epochs $N_{epoch}$, margin value $\alpha$, Adam hyperparameters %$\gamma$, $\beta_1$, $\beta_2$
%    \State Initialize parameters $\theta$ for ${\mathcal{V}}$,  $\mathcal{T}$, $f_t^{img}$, %$f_v^{img}$, $f_t^{txt}$, $f_v^{txt}$
%    \For{$i = 1 \to N_{epoch}$}
%        \For{$j = 1 \to N_{batch}$}
%            \State $\mathcal{L}_v = 0$, $\mathcal{L}_t = 0$
%            \For{$(x_v, x_t)$ in ${batch}_j$}
%                \State Select $N_v$, $N_t$ from ${batch}_j$
%                \State $x_{v_1}, x_{v_2} = \mathcal{A}(x_v)$
%                \State $v_1 = \mathcal{V}(x_{v_1}), v_2 = \mathcal{V}(x_{v_2})$
%                \State $t = \mathcal{T}(x_t)$
%                \State $\mathcal{L}_{intra} += infoNCE(v_1, v_2, N_v)$
%                \State $\mathcal{L}_{inter} += PwR(v_1, t, N_t) + PwR(v_2, t, N_t)$
%            \EndFor
%            \State $\mathcal{L} = \mathcal{L}_{intra} + \mathcal{L}_{inter}$
%        	\State $\theta \gets Adam(\nabla_{\theta} \mathcal{L}, \theta, \alpha, \gamma, \beta_1, %\beta_2)$
%        \EndFor
%    \EndFor
%\end{algorithmic}  
%\end{algorithm}

\section{Experiments}
\label{sec:experiments}
According to the conventional training procedure of self-supervised system, the following evaluation is two-stage, including the pre-training stage and the feature-based fine-tuning stage.

\begin{table*}[t]
\centering
\small
 \begin{tabular}{lcccc|cccc} \toprule
& \multicolumn{4}{c|}{\textbf{TVDIM\_SMALL}} & \multicolumn{4}{c}{\textbf{TVDIM\_LARGE}}\\ 
\cmidrule{2-5} \cmidrule{6-9}
& \multicolumn{2}{c}{\textbf{0.8M}} & \multicolumn{2}{c|}{\textbf{3.3M}} & \multicolumn{2}{c}{\textbf{0.8M}} & \multicolumn{2}{c}{\textbf{3.3M}}\\
\cmidrule{2-5} \cmidrule{6-9}
Model & AMDIM & TVDIM & AMDIM & TVDIM & AMDIM & TVDIM & AMDIM & TVDIM \\ 
 & (Lin., MLP) & (Lin., MLP) & (Lin., MLP) & (Lin., MLP) & (Lin., MLP) & (Lin., MLP) & (Lin., MLP) & (Lin., MLP) \\ 
\midrule
CIFAR10 & 66.4, 71.7 & 68.7, 72.3 & 66.9, 71.8 & 70.5, 73.4 & 69.6, 73.1 & 70.5, 73.8 & 71.4, 74.8 & 72.2, 75.0\\
CIFAR100 & 43.1, 45.9 & 45.0, 47.5 & 45.2, 48.2 & 47.6, 49.8 & 45.0, 47.3 & 48.3, 49.7 & 46.6, 49.9 & 50.0, 51.3\\
STL10 & 66.8, 70.1 & 70.1, 71.9 & 68.6, 71.8 & 71.7, 73.5 & 68.2, 72.1 & 72.9, 74.1 & 68.5, 74.1 & 73.2, 75.5 \\
%ImageNet & --.-, --.- & --.-, --.- & --.-, --.- & --.-, --.- & --.-, --.- & --.-, --.- & --.-, --.- & --.-, --.- \\
\bottomrule
 \end{tabular}
 \caption{Top-1 classification accuracy (\%) of various image classification tasks. We adopt both linear layer ($Lin.$) and MLP layer ($MLP$) as the classification layer to evaluate the performance. Noted that the pretraining setting and fine-tuning setting are different from that of AMDIM. Therefore, the results shown here are not comparable with the results presented in the AMDIM paper. AMDIM unifies the training dataset for both pretraining stage and fine-tuning stage, and their Linear layer and MLP layer are also fine-tuned without gradient back-propagation to image encoder during pretraining stage. But these settings is not applicable to our case. Here we consider a more challenge setting that image encoder is pretrained from a unified large-scale dataset, and then applied to various downstream tasks.}
 \label{table:acc_overall}
\end{table*}

\subsection{Pre-Training}
For better evaluation, We use two different image-text datasets to do the self-supervised pretraining. For the selection of dataset, there are two prerequisites: obtaining without human annotation and the high information overlap within each image-text pair. Two datasets shown below have different data scale and different average lengths of text sequence. Below presents the dataset descriptions and the implementation details.

\subsubsection{Conceptual Captions Dataset}

Conceptual Captions\cite{sharma2018conceptual} is a collection of 3.3 million image-caption pairs automatically scraped from alt-text enabled web images. The average length of captions is around 10. It is easy to collect, and each image-text pair exists high information overlap. However, the automatic collection and sanitation process leaves some \textbf{noise} and the captions are sometimes not human-like or short on details. Following the design standard of AMDIM, we use two network structures with different parameter scales,

\noindent $\mathrm{TVDIM}_{small}$: (ndf=64, nrkhs=512, ndepth=8)\footnote{$rkhs$ is the abbreviation of Reproducing Kernel Hilbert Spaces, which denotes the dimension of visual representation.},

\noindent $\mathrm{TVDIM}_{large}$: (ndf=96, nrkhs=768, ndepth=8).

\noindent The pretraining process is on 4 P100 GPUs with a total batch size of 480 for 15 epochs. The negative samples for each processing image-text pair are selected from the current batch it belongs to. We use the Adam optimizer \cite{kingma2014adam} with initial learning rates $\gamma$ of 1e-4, $\beta_1=0.9$, $\beta_2=0.999$, weight decay of 1e-4. The margin value $\alpha$ of pairwise ranking loss is set as 0.5. In terms of data augmentation, we apply resized cropping, color jitter, and random conversion to grayscale for each image input.

\subsubsection{MM-IMDB Dataset}
MM-IMDb dataset \cite{arevalo2017gated} is built from the Movielens 20M dataset\footnote{http://grouplens.org/datasets/movielens/}. The MM-IMDb dataset comprises 25,959 movies along with their plot, poster, genres and other 50 additional metadata fields. In our setting, we use the (poster, plot) as our image-text pair. Note that each plot contains on average 92.5 words, which is quite longer than that of Conceptual Captions Dataset. Considering its small scale, we adopt a relatively small-scale image encoder structure:

\noindent $\mathrm{TVDIM}_{MM-IMDb}$: (ndf=64, nrkhs=512, ndepth=1).

\noindent We set the batch size as 256 and pre-train for 80 epochs. The negative samples are also selected from training batches. The optimizer has the same setting as Conceptual Captions dataset, except for the learning rate $\gamma$ of 5e-4. The margin value $\alpha$ is 0.5 as well.

\begin{table}[t]
\centering
 \begin{tabular}{lcccc} \toprule
& \multicolumn{4}{c}{\textbf{F-Score}} \\ 
\cmidrule{2-5}
Model & sample & micro & macro & weight  \\ \midrule
CNN  & 35.0 & 34.0 & 21.0 & 37.0 \\
Supervised L. & 37.2 & 37.4 & 21.2 & 36.8 \\
AMDIM & 37.7 & 37.7 & 20.7 & 37.4 \\
TVDIM & \textbf{39.8} & \textbf{39.5} & \textbf{24.1} & \textbf{39.5} \\
\bottomrule
 \end{tabular}
 \caption{Comparison results of MM-IMDb dataset. The MM-IMDb training dataset is applied for both the pretraining stage and the fine-tuning stage. \textbf{CNN} denotes the model from Arevalo et al. [2017].}
 \label{table:MM-IMDb}
\end{table}

% \footnotesize{\cite{arevalo2017gated}}

\subsection{Fine-Tuning on Downstream Tasks}

%Our goal is to learn the better visual representations with the guidance from textual information. 
Following the evaluation protocol of previous works \cite{kolesnikov2019revisiting,bachman2019learning}, we evaluate TVDIM on several image classification benchmarks, including CIFAR10, CIFAR100 and STL10. The input images are uniformly resized to 128*128 before encoding. Similar to the conventional setting, 
the self-supervised pretrained image encoder (from training dataset $D_{ssl}$) is frozen and directly used to generate visual representations, and then we train a linear classifier or a MLP classifier on the downstream image classification dataset $D_{cl}$. Unlike previous settings, two training datasets $D_{ssl}$, $D_{cl}$ are not the same one: $D_{ssl}$ is uniformly the Conceptual Captions dataset. Therefore, our evaluation method belongs to the transfer task that is also widely used for evaluating self-supervised visual representation learning \cite{bachman2019learning}. The comparison results are presented in Table \ref{table:acc_overall}. Besides the different scales of image encoder, we also compare the performance from two different sizes of self-supervised training dataset. Here we assign the full Conceptual Captions dataset as the big one \textbf{3.3M}, and randomly select a quarter of it as the small one \textbf{0.8M}. It is obvious that TVDIM outperforms AMDIM on various image classification benchmarks.

Besides that, we also perform the comparison experiments on the multi-modal dataset MM-IMDb, where $D_{ssl}$ and $D_{cl}$ is unified into the same dataset. The downstream task is still the image classification task. The performance is shown in Table \ref{table:MM-IMDb}. We present the result of supervised learning of our image encoder, which achieves slightly better performance improvement compared with the previously-published result. Then, we compare the performance of AMDIM and TVDIM, and TVDIM yields the significant improvement with the help of inter-modality learning.

To sum up, the proposed TVDIM is able to effectively encode the image-related text information into visual representations, and further promote the visual understanding ability of machine. More than that, image-text pairs are accessible and massive without human annotation, which is crucial for self-supervised learning.

\subsection{Analysis}

\subsubsection{Ablation Studies}
\label{subsec:ablation}
We conduct ablation studies from three aspects in Table \ref{table:Ablation}. CIFAR10 and Place205 are selected as the representatives for the transfer setting. For MM-IMDb, we only perform the F1-macro score because it is the most widely-used F1 score. First, we can see that batch normalization is vital to the calculation of inter-modality mutual information. Second, with the assistance of visual local features, the inter-modality mutual information maximization achieves better performance. Finally, we find that the performance get further improved by maximizing mutual information in both image feature space and text feature space.

\begin{table}[t]
\centering
 \begin{tabular}{lccc} \toprule
Model & CIFAR10 & STL10 & MM-IMDb  \\ 
 & Accuray(\%) & Accuray(\%) & F1-macro  \\ 
\midrule
TVDIM & \textbf{72.2} & \textbf{73.2} & \textbf{24.1} \\
\cmidrule{2-4}
\ \ \ $w/o$ BN  & 69.5 & 70.1 & 22.1  \\
\ \ \ $w/o$ Local & 70.7 & 70.8 & 22.8  \\
\ \ \ $w/o$ V2T & 71.2 & 71.9 & 23.5  \\
\bottomrule
 \end{tabular}
 \caption{The comparison results of ablation study. Here $BN$ denotes batch normalization; $Local$ denotes that the local visual representations are considered for inter-modality learning; $V2T$ denotes the inter-modality mutual information maximization in text feature space. The results of CIFAR10 and STL10 are from $TVDIM_{large}$ on 3.3M dataset.}
 \label{table:Ablation}
\end{table}

\subsubsection{Analysis of Pairwise Ranking Loss}
Considering the information gap between vision modality and text modality, we adopt pairwise ranking loss for inter-modality self-supervised learning. Table \ref{table:Loss Function} proves the superiority of pairwise ranking loss in this case.
According to the definitions, NCE is from the angle of probability, which assigns infinite reward to positive samples and infinite penalty to negative samples. It is reasonable for our visual intra-modality learning, because global features and local features are derived from the same image. For inter-modality, image-text pairs have high information overlap while the information gap also exists. Therefore, if we infinitely maximize their mutual information, it would run counter to our desire. Pairwise ranking loss is defined from the difference of information amount. More specifically, the boundary of reward and penalty is that the difference reaches the requirement of margin.

\begin{table}[t]
\centering
 \begin{tabular}{lccc} \toprule
Model & CIFAR10 & STL10 & MM-IMDb  \\ 
 & Accuray(\%) & Accuray(\%) & F1-macro  \\ 
\midrule
NCE  & 67.7 & 69.8 & 22.0  \\
Pairwise R. & \textbf{72.2} & \textbf{73.2} & \textbf{24.1} \\
\bottomrule
 \end{tabular}
 \caption{Comparison results of different contrastive learning method for inter-modality learning. The results of CIFAR10 and STL10 are from $TVDIM_{large}$ on 3.3M dataset.}
 \label{table:Loss Function}
\end{table}

\subsubsection{Impact of Text Encoder}
\label{subsec:test_encoder}
We have tried several different settings for text encoder. Here we consider MM-IMDb dataset to evaluate the influence of text encoder $\mathcal{T}(*)$. Compared with Conceptual Captions dataset, the text descriptions of MM-IMDb dataset is more complex; Therefore, the impact of text encoder is more obvious.
Table \ref{table:Text Encoder} illustrates that the mean of word embedding\footnote{https://drive.google.com/file/d/0B7XkCwpI5KDYNlNUTTlSS21pQmM/edit} achieves the best performance, even though it is the simplest way. Our explanation is that the trainable parameters added from text encoder (CNN or LSTM) reduce the difficulty of self-supervised learning. To be specific, it becomes relatively easier to maximize the mutual information between image and text. Consequently, the performance goes down. We also have tried the contextualized word embedding from BERT \cite{devlin2018bert}, but it still obtains the suboptimal performance. It proves that contextualized word embeddings should be used via fining-tuning BERT model rather than directly used. 
%Even though the mean of word embedding has the best performance here, it still has the shortage of poor diversity. We will continue to find the better solutions for text encoder in future work.
%ay not be the best way to represent text information since it is lack of diversity. We leave this to future work.

\begin{table}[t]
\centering
 \begin{tabular}{lcccc} \toprule
& \multicolumn{4}{c}{\textbf{F-Score}} \\ 
\cmidrule{2-5}
Text Encoder & sample & micro & macro & weight  \\ \midrule
CNN & 38.2 & 38.2 & 23.0 & 38.5 \\
LSTM & 38.5 & 38.4 & 22.4 & 38.5 \\
BERT_mean & 39.6 & 39.2 & 23.5 & 39.2 \\
W2V_mean & \textbf{39.8} & \textbf{39.5} & \textbf{24.1} & \textbf{39.5} \\
\bottomrule
 \end{tabular}
 \caption{Analysis of the selection of text encoder on MM-IMDb dataset. $W2V\_mean$ and $BERT\_mean$ denote the mean operation on the pretrained word embedding, and word embedding is frozen during self-supervised learning.}
 \label{table:Text Encoder}
\end{table}

\subsection{Discussion}
We have shown that pairwise ranking loss is an effective contrastive learning loss for self-supervised learning. However, when we attempt to adopt it to the visual intra-modality learning, the performance is quite poor. We have checked the intermediate results of the pretraining stage and find that the loss value drops sharply, which means that it is quite easy to convergence. In other words, the learning process of self-supervised pretraining is lack of difficulty, so our image encoder cannot learn sufficient knowledge. We conclude that it is crucial to control the difficulty of self-supervised learning in an appropriate range. Within this range, the increase of difficulty is beneficial for more robust visual representations. This point of view is also reflected by the impact of text encoder in Section \ref{subsec:test_encoder}. Without introducing extra trainable parameters into TVDIM, the mean of word embeddings achieves the best performance. But, for the design of text encoder, we need to acknowledge that the mean of word embeddings is just the better one in our current solutions but not the best, it still has a shortage of poor diversity. We will continue to find better solutions for text encoder in future work.

\section{Conclusion}
\label{conclusion}
In this paper, we present a novel self-supervised visual representation learning method by considering both visual intra-modality learning and vision-text inter-modality learning. Various media provide massive image-text pairs that exist high information overlap and are easily obtained without human annotations. The proposed TVDIM is able to effectively encode valuable information from text modality into visual representations. Considering the existence of information gap, we adopt pairwise ranking loss to maximize the mutual information for inter-modality learning. Theoretically, TVDIM is a model-agnostic technique that can be applied to any existing self-supervised visual representation learning frameworks regardless of network structures. Experimental results show that TVDIM is the better self-supervised strategy for visual representation learning and significantly outperforms previous methods.


\bibliographystyle{named}
\bibliography{TVDIM}

\end{document}